\documentclass{article}
\usepackage[utf8]{inputenc}
\usepackage{natbib}
\usepackage{fancyhdr, fullpage, parskip,amsfonts}
\usepackage{times,latexsym}
\usepackage[T1]{fontenc}
\usepackage{algorithm,algorithmic}

\makeatletter
\long\def\@makecaption#1#2{
  \vskip 0.8ex
  \setbox\@tempboxa\hbox{\small {\bf #1:} #2}
  \parindent 1.5em  %
  \dimen0=\hsize
  \advance\dimen0 by -3em
  \ifdim \wd\@tempboxa >\dimen0
  \hbox to \hsize{
    \parindent 0em
    \hfil 
    \parbox{\dimen0}{\def\baselinestretch{0.96}\small
      {\bf #1.} #2
    } 
    \hfil}
  \else \hbox to \hsize{\hfil \box\@tempboxa \hfil}
  \fi
}
\makeatother

\usepackage{microtype}
\usepackage{graphicx}
\usepackage{subfigure}
\usepackage{booktabs} %
\usepackage{enumitem,multirow,adjustbox}
\usepackage{arydshln} %

\usepackage{hyperref}
\usepackage{url}
\usepackage{xcolor}		%
\definecolor{darkblue}{rgb}{0, 0, 0.5}
\definecolor{beaublue}{rgb}{0.74, 0.83, 0.9}
\definecolor{gainsboro}{rgb}{0.86, 0.86, 0.86}
\definecolor{kleinblue}{rgb}{0,0.18,0.65}
\hypersetup{colorlinks=true,citecolor=kleinblue, linkcolor=kleinblue, urlcolor=kleinblue}

\usepackage{amsmath,amsfonts,bm}

\def\eqref#1{equation~\ref{#1}}

\def\1{\bm{1}}

\DeclareMathAlphabet{\mathsfit}{\encodingdefault}{\sfdefault}{m}{sl}
\SetMathAlphabet{\mathsfit}{bold}{\encodingdefault}{\sfdefault}{bx}{n}

\def\gF{{\mathcal{F}}}

\def\gN{{\mathcal{N}}}

\def\gP{{\mathcal{P}}}

\def\gX{{\mathcal{X}}}

\newcommand{\E}{\mathbb{E}}

\newcommand{\R}{\mathbb{R}}

\usepackage{amsmath}
\usepackage{amssymb}
\usepackage{mathtools}
\usepackage{amsthm}

\usepackage[capitalize,noabbrev]{cleveref}

\theoremstyle{plain}
\newtheorem{theorem}{Theorem}[section]

\theoremstyle{definition}
\newtheorem{definition}[theorem]{Definition}

\theoremstyle{remark}

\usepackage{xspace}

\newcommand{\dataset}{{\cal D}}

\newenvironment{myquotation}{\setlength{\leftmargini}{0em}\quotation}{\endquotation}

\definecolor{purple}{HTML}{000000}

\newcommand{\customfootnotetext}[2]{{%
  \renewcommand{\thefootnote}{#1}%
  \footnotetext[0]{#2}}}%

\begin{document}

\title{Positional Information Matters for Invariant In-Context Learning: \\A Case Study of Simple Function Classes}

\author{
    Yongqiang Chen$^{*\dag}$
    \and
    Binghui Xie$^{\dag}$
    \and
    Kaiwen Zhou$^{\dag}$
    \and
    Bo Han$^\diamond$
    \and
    Yatao Bian$^{\ddag}$
    \and
    James Cheng${^\dag}$
}

\date{}

\maketitle
\customfootnotetext{$*$}{Work done during an internship at Tencent AI Lab.}

\customfootnotetext{$\dag$}
{The Chinese University of Hong Kong; e-mail: \text{\{yqchen,bhxie21,kwzhou,jcheng\}@cse.cuhk.edu.hk}}
\customfootnotetext{$\diamond$}
{Hong Kong Baptist University; e-mail: \text{bhanml@comp.hkbu.edu.hk}}
\customfootnotetext{$\ddag$}
{Tencent AI Lab; e-mail: \text{yatao.bian@gmail.com}}

\maketitle

\begin{abstract}
    In-context learning (ICL) refers to the ability of a model to condition on a few in-context demonstrations (input-output examples of the underlying task) to generate the answer for a new query input, without updating parameters.
    Despite the impressive ICL ability of LLMs, it has also been found that ICL in LLMs is sensitive to input demonstrations and limited to short context lengths.
    To understand the limitations and principles for successful ICL, we conduct an investigation with ICL linear regression of transformers.
    We characterize several Out-of-Distribution (OOD) cases for ICL inspired by realistic LLM ICL failures and compare transformers with DeepSet, a simple yet powerful architecture for ICL.
    Surprisingly, DeepSet outperforms transformers across a variety of distribution shifts, implying that preserving permutation invariance symmetry to input demonstrations is crucial for OOD ICL.
    The phenomenon specifies a fundamental requirement by ICL, which we termed as \textit{ICL invariance}.
    Nevertheless, the positional encodings in LLMs would break  ICL invariance.
    To this end, we further evaluate transformers with identical positional encodings and find that  preserving ICL invariance in transformers achieves state-of-the-art performance across various ICL distribution shifts.
\end{abstract}

\section{Introduction}
\begin{figure}[t]
    \centering
    \includegraphics[width=0.9\textwidth]{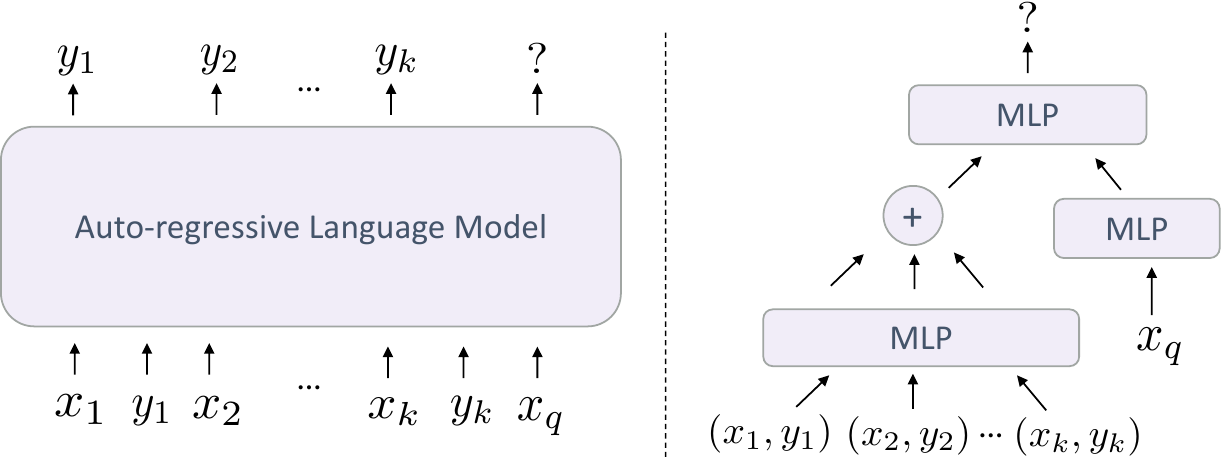}
    \caption{Illustration of linear regression ICL with auto-regressive Transformer (left) and DeepSet (right).
        Given $k$ input demonstrations $(x_1,y_1),(x_2,y_2),...,(x_k,y_k)$, and the query input $x_k$, Transformer adheres to the paradigm of the auto-regressive language model to infer the labels in an auto-regressive manner. In contrast, DeepSet jointly models the $k$ sequential demonstrations as a set, and produces the output of the query based on the set-aggregated representations.
    }
    \label{fig:arch_illustration}
\end{figure}

Transformers~\citep{transformers} have been the \textit{de facto} building block of foundational models for a variety of applications such as computer vision~\citep{vit,clip,flamingo}, natural language processing~\citep{bert,gpt3,llama}.
By training on massive text corpora to simply predict the next word given the prefixing words, transformer-based large language models (LLMs) are emerging to do in-context learning (ICL)~\citep{gpt3}.
ICL refers to the ability of a model to solve new tasks conditioning on a few demonstrations, while without updating their parameters.
Despite the success of LLMs in ICL a variety of tasks~\citep{gpt3}, it has also been found that ICL in LLMs is sensitive to the \textit{distribution shifts} in input demonstrations~\citep{compositional_gap,label_important,label_bias,irrelevant_text,icl_shortcut,underspecific_icl,chatgpt_ood} and the lengths of the context windows~\citep{context_window,length_gen}.
However, there is little understanding of the fundamental limitations and principles for successful ICL, which raises a curious research question:
\begin{myquotation}\centering
    \textit{What fundamental properties do we need for successful ICL under distribution shifts?}
\end{myquotation}
To answer the question, in this paper, we conduct an investigation with ICL linear regression following the setting of~\cite{ICL_simple_function,ICL_MLP}.
We derive the distribution shifts for ICL linear regression inspired by the LLM ICL failures: i) Generalizing to out-of-distribution (OOD) demonstrations~\citep{compositional_gap,label_bias,irrelevant_text,icl_shortcut,chatgpt_ood}, and ii) Generalizing to OOD context lengths~\citep{context_window,length_gen}.
Following~\cite{ICL_MLP}, we compare the transformer with DeepSet, a set-based MLPs~\citep{deep_set,set_function}.
Specifically, the Transformer is an auto-regressive architecture used by the state-of-the-art LLMs~\citep{gpt3}, which infers the labels of new inputs based on the prefix input-output pairs sequentially, shown as in Fig.~\ref{fig:arch_illustration}.
In contrast, DeepSet models the input demonstrations as a set and conditions on the aggregated demonstration representations to produce the answer to the query.

When there are no distribution shifts, both DeepSet and Transformer exhibit the expected ICL capabilities, as also observed by~\cite{ICL_MLP}. Surprisingly, when distribution shifts occur, DeepSet can outperform the Transformer despite its simple architecture.
Since the key difference between DeepSet and Transformer lies only in the modeling of input demonstrations, we conclude the permutation invariance symmetry modeled within DeepSet but not in the Transformer is crucial for ICL, especially under distribution shifts. We term the symmetry as \textit{ICL invariance}.
The auto-regressive manner in Transformer will however break the desired ICL invariance.
To further verify our findings, we design a new Transformer architecture with identical positional encodings, which commits to ICL invariance. The experimental results show that the new Transformer architecture outperforms both DeepSet and Transformer across various distribution shifts, serving as a piece of strong evidence for the requirement of ICL invariance.

\section{Problem Formulation}
We begin by formalizing our problem settings following~\cite{ICL_simple_function,ICL_MLP}, and then define the distribution shifts of ICL.
\subsection{ICL linear regression}
\label{sec:icl_lin}
\paragraph{Data.}
We denote inputs and labels as $x_i\in\R^d$ and $y\in\R$, respectively.
The ICL is to predict the label $y_q$ for the query input $x_q$, conditioning on $k$ demonstrations $(x_1,y_1),(x_2,y_2),...,(x_k,y_k)$, which compose
a prompt $p=\{(x_i,y_i)\}_{i=1}^k$.
The prompt is generated as follows
\begin{enumerate}[label=(\alph*),leftmargin=*]
    \item Draw $x_1,...,x_k$ independently from a distribution $\dataset_\gX$;
    \item Sample the underlying function $f\in\gF$ from $\dataset_\gF$ over the function class $\gF$. For linear regression, $f$ is the linear function $f(x)=w^Tx$ with the weight $w\sim\gN(0,I_d)$, where $I_d$ is the identity matrix in $d$ dimensions;
    \item Generating the prompt with $k$ demonstrations as $p=(x_1,f(x_1),x_2,f(x_2),...,x_k,f(x_k))$;
\end{enumerate}
The prompt along with the query example is called the prompt prefix $p^i=(x_1,f(x_1),x_2,f(x_2),...,x_i,f(x_i),x_{i+1})$ with $i$ demonstrations and $(i+1)$-th example as the query example.
Essentially, the aforementioned data model specifies the distribution of prompt prefixes, which can be considered as the inputs to the model in ICL.
Then, we need to train a model $M_\theta$ parameterized by $\theta$ that aims to minimize the expected loss over all prompt prefixes:
\begin{align}\label{eq:icl_loss}
    \min_\theta \E_p[\frac{1}{k+1}\sum_{i=0}^kl(M_\theta(p^i),f(x_{i+1}))],
\end{align}
where $l(\cdot,\cdot)$ is the loss function. In the case of linear regression, we consider the MSE loss.

\paragraph{Architecture.} We consider two model architectures for implementing $M$ following~\cite{ICL_MLP}. For the Transformer, we use the same architecture as in~\cite{ICL_simple_function} from the GPT-2 family~\cite{gpt2}, which consists of $12$ layers, $8$ attention heads and a hidden dimension of $256$. The Transformer architecture generates the next word prediction in an auto-regressive manner and naturally fits into the prompt sequence.

Besides, we consider DeepSet~\citep{deep_set}, which is a set-based MLP~\citep{set_function} that is also leveraged by~\cite{ICL_MLP}. The DeepSet architecture predicts the output of the query example via $\psi(\rho(\frac{1}{k}\sum_{i=1}^k\phi(x_i,y_i)),x_{k+1})$, where $\phi,\rho,\psi$ are all MLPs that encode the input demonstrations, the prompt, and the prompt prefix, respectively. The hidden dimension is set to be $500$ with a depth of $5$, which corresponds to the ``small'' variant of~\cite{ICL_MLP}.
Although \cite{ICL_MLP} also adopt the same DeepSet architecture, they focus on revealing the advances of Transformer over DeepSet under distribution shifts. In contrast, we find DeepSet can outperform Transformer due to its modeling of ICL invariance.

\section{In-Context Learning Invariance}
It has been proved by~\cite{ICL_MLP} that the optimal model under the setting of Sec.~\ref{sec:icl_lin} emulates the ordinary least squares (OLS) or the ridge regression on the support of the training distribution, aligning with existing empirical observations by~\cite{ICL_simple_function}.
The natural question is then, what would guarantee the performance outside the support of the training distribution?

To answer the question,
We consider three types of distribution shifts, ranging from the demonstration inputs $\{x_i\}_{i=1}^k$, demonstration labels $\{y_i\}_{i=1}^k$, and the length of context $k$:
\begin{itemize}[leftmargin=*]
    \item Distribution shifts in $\{x_i\}_{i=1}^k$~\citep{icl_shortcut,irrelevant_text,chatgpt_ood}: We simulate the OOD demonstrations inputs by changing $\dataset_\gX=\gN(\mu\cdot\mathbf{1},I_d)$ across training and testing. During training, $\mu=0$ while $\mu$ is changed to be $2,4$ during testing.
    \item Distribution shifts in $\{y_i\}_{i=1}^k$~\citep{label_bias,label_bias2,label_bias3}: We impose distribution shifts to labels by injecting label noises to generate $y_i=f(x_i)+\sigma \eta$, with $\eta\sim\gN(0,1)$. During training, $\sigma=0$ while $\sigma$ is changed to be $1$ during testing.
    \item Distribution shifts in $k$~\citep{length_gen,context_window}: To simulate distribution shifts in context lengths, we follow~\cite{ICL_simple_function} that adopt a context window $k\in[0,50]$ using the scheme of Curriculum learning~\citep{curr_learn}, while we adopt $k$ up to $100$ in testing.
\end{itemize}

\begin{figure}[ht]
    \vspace{-0.2in}
    \centering
    \subfigure[OOD ICL with $\mu=0$.]{
        \includegraphics[width=0.31\textwidth]{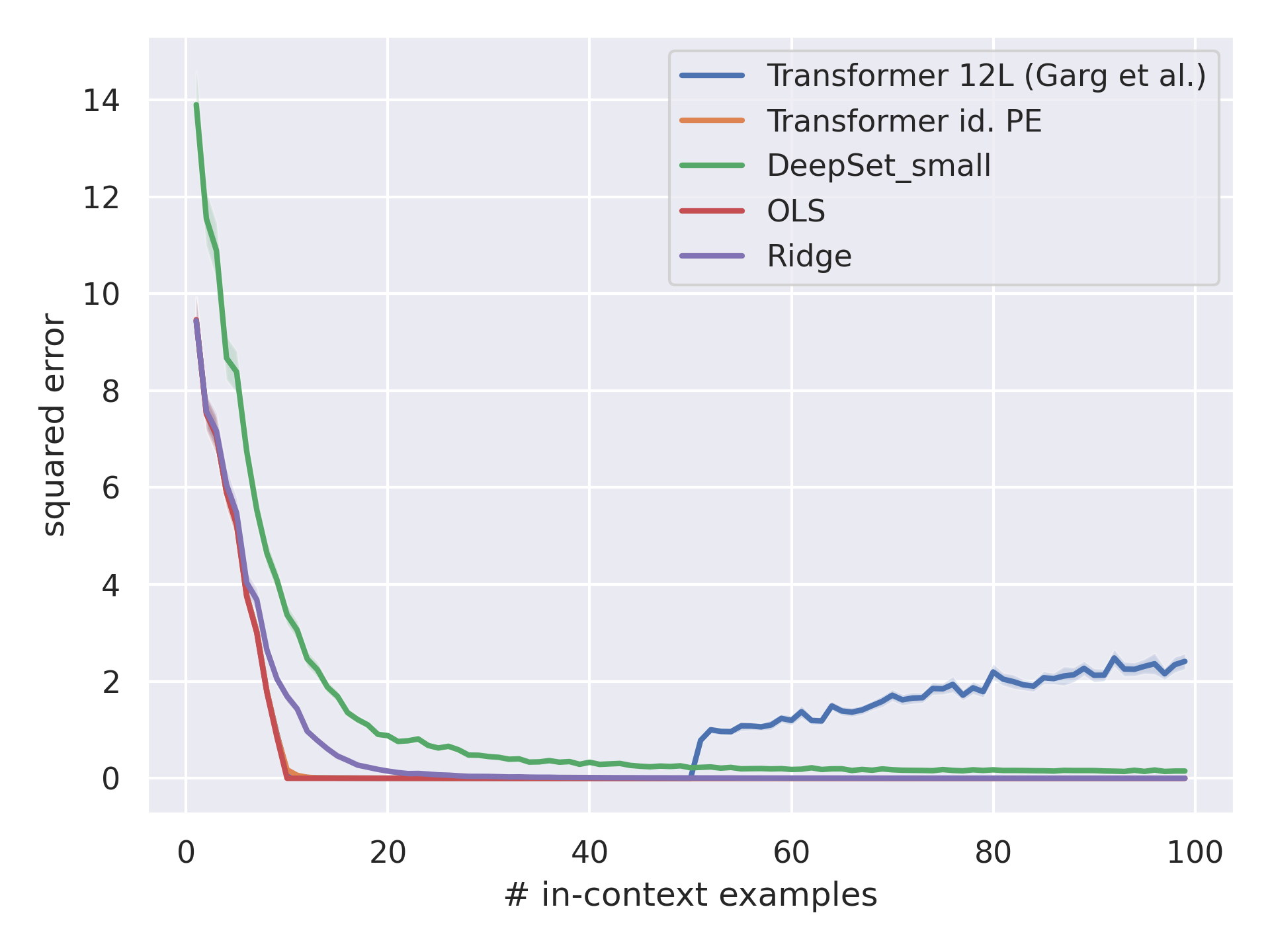}
        \label{fig:label_noise=False_mu=0}
    }
    \subfigure[OOD ICL with $\mu=2$.]{
        \includegraphics[width=0.31\textwidth]{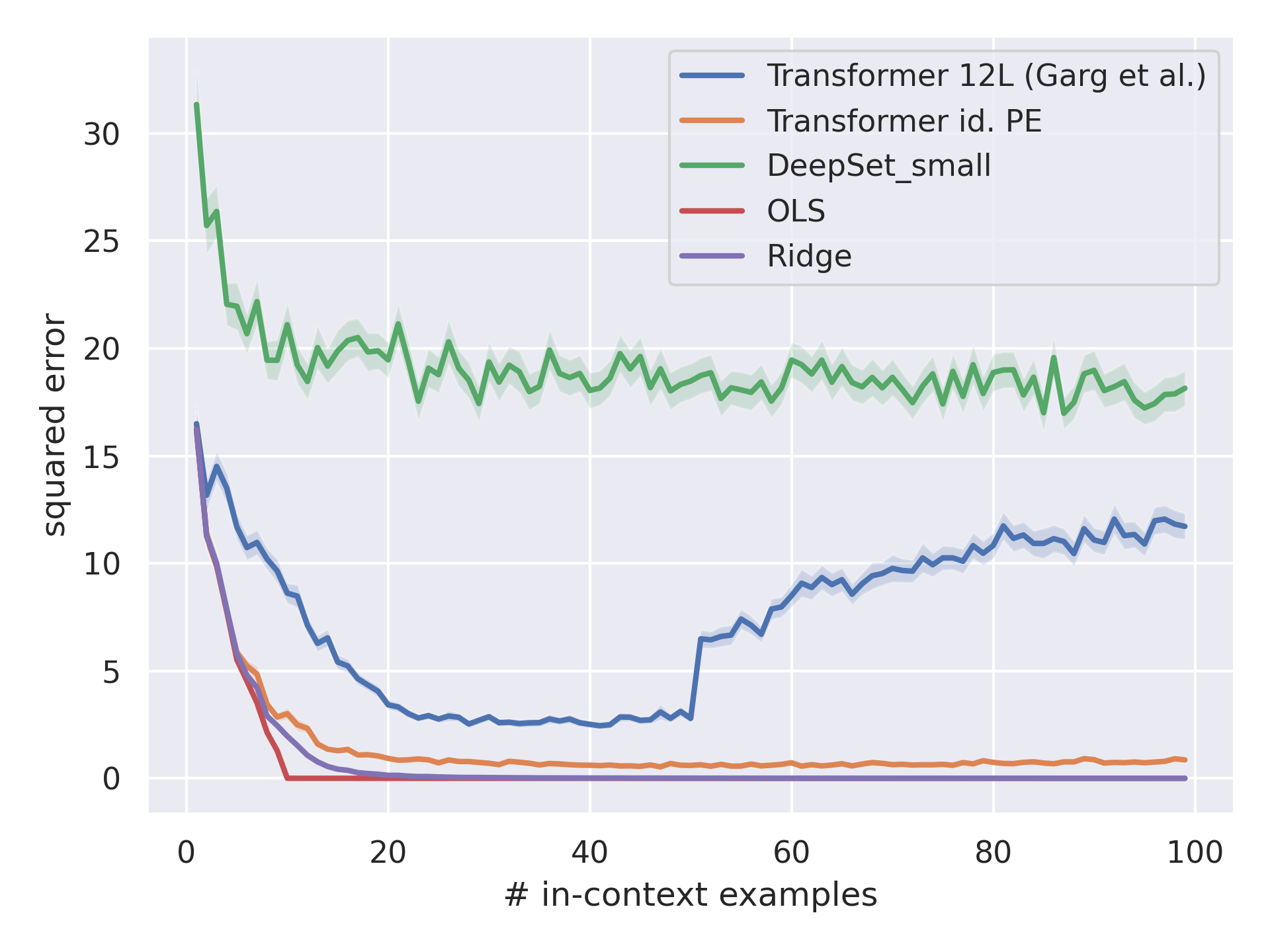}
        \label{fig:label_noise=False_mu=2}
    }
    \subfigure[OOD ICL with $\mu=4$.]{
        \includegraphics[width=0.31\textwidth]{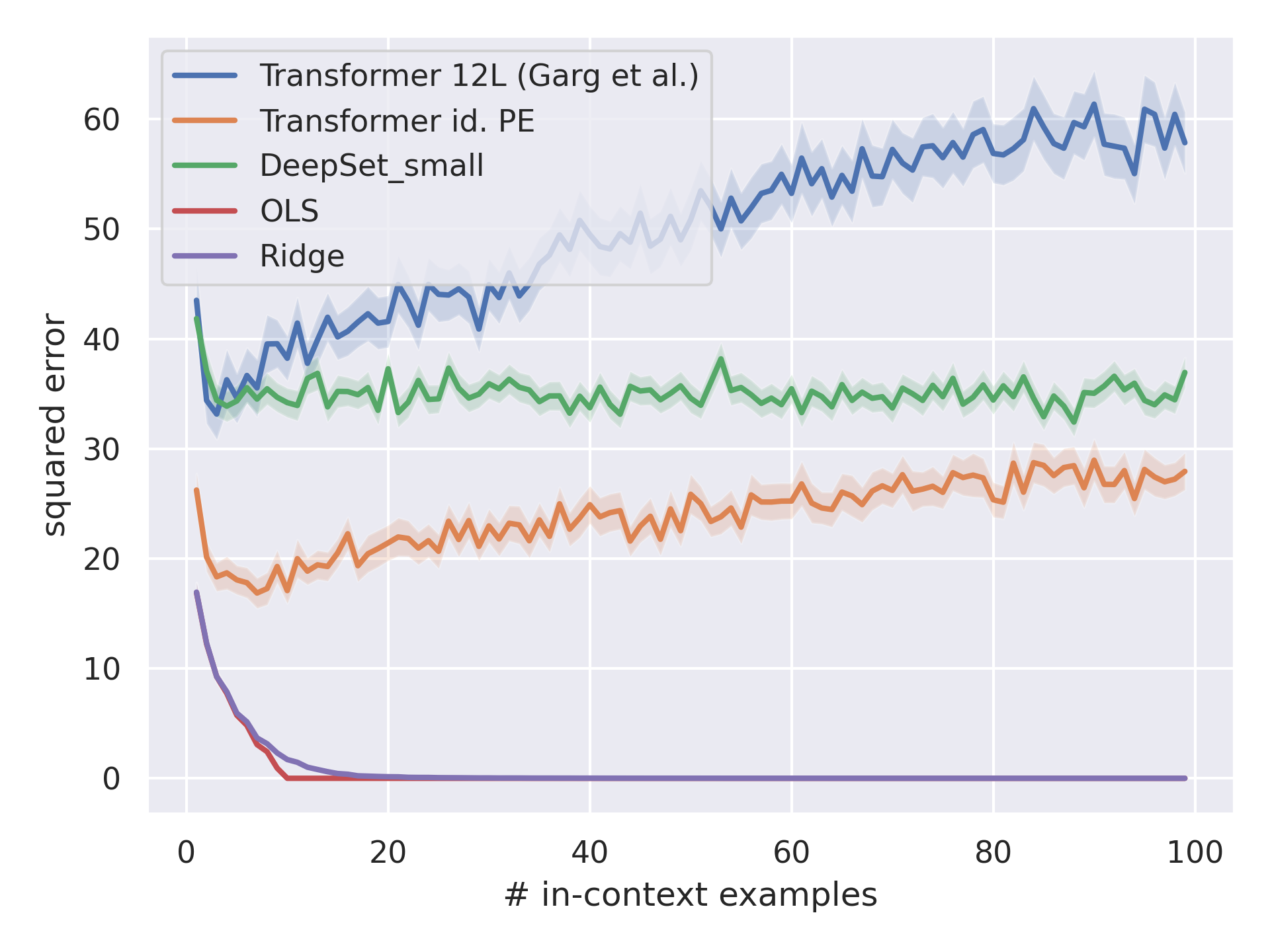}
        \label{fig:label_noise=False_mu=4}
    }
    \subfigure[OOD ICL with $\mu=0$ and $\sigma=1$.]{
        \includegraphics[width=0.31\textwidth]{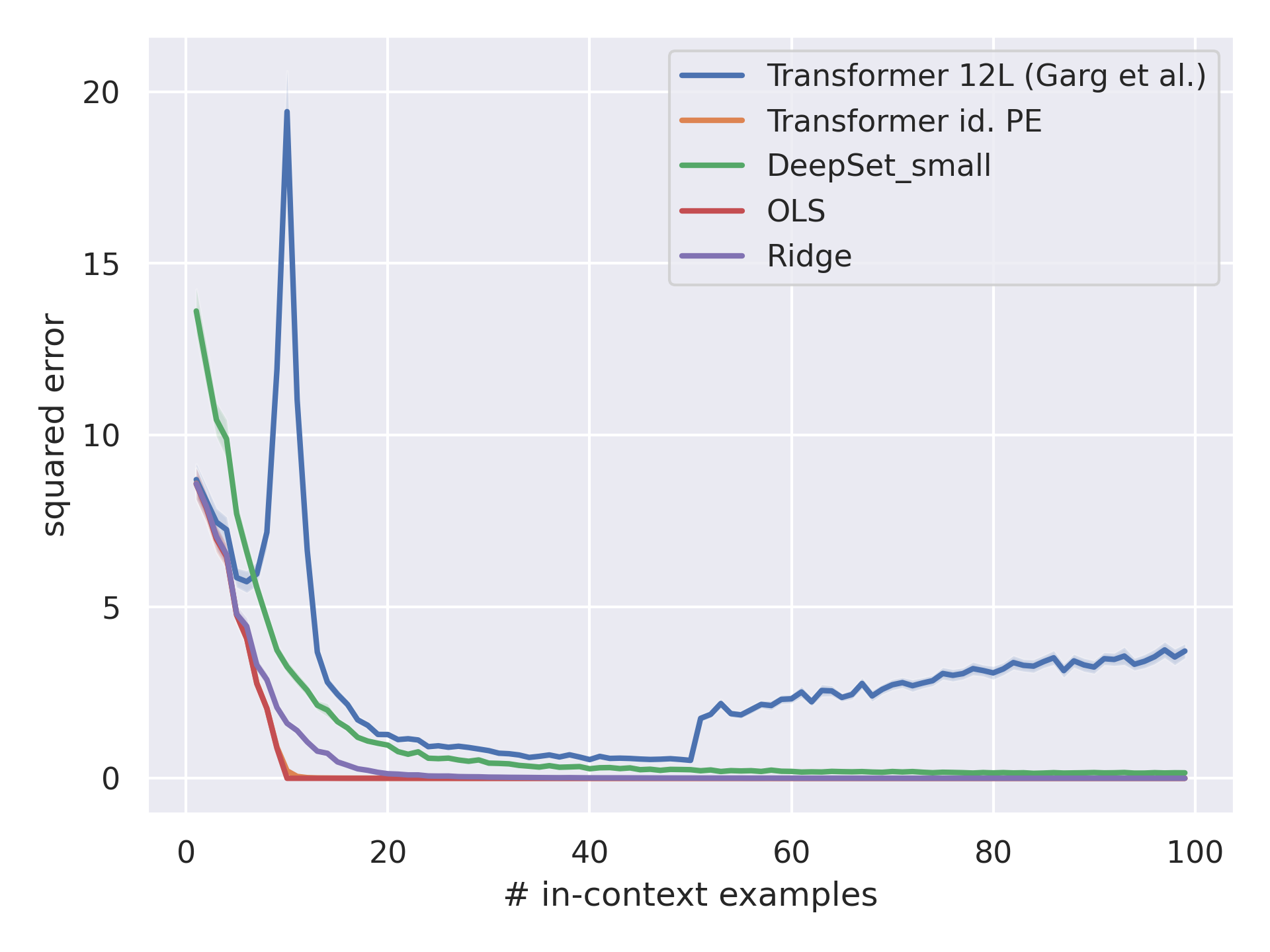}
        \label{fig:label_noise=True_mu=0}
    }
    \subfigure[OOD ICL with $\mu=2$ and $\sigma=1$.]{
        \includegraphics[width=0.31\textwidth]{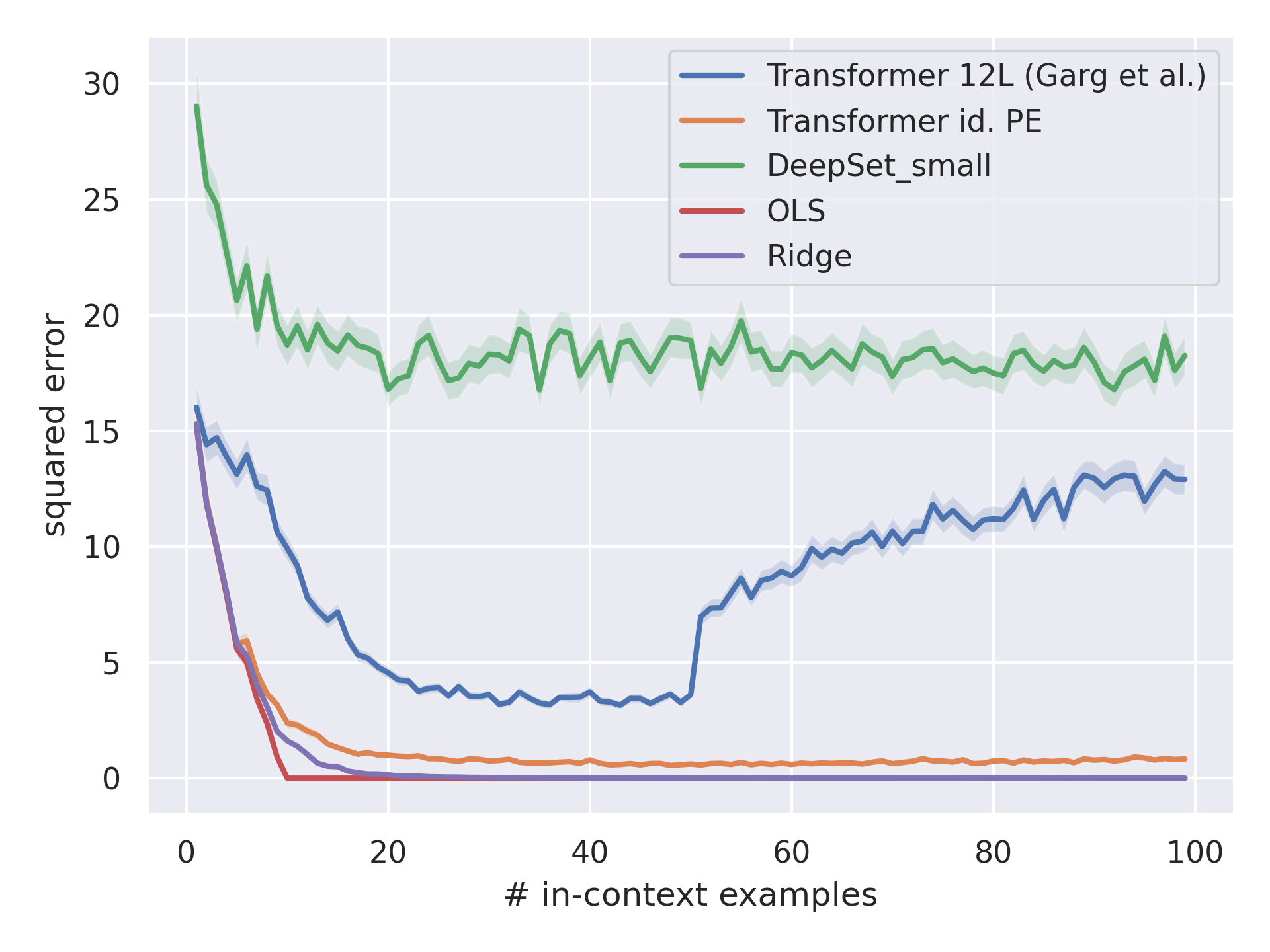}
        \label{fig:label_noise=True_mu=2}
    }
    \subfigure[OOD ICL with $\mu=4$ and $\sigma=1$.]{
        \includegraphics[width=0.31\textwidth]{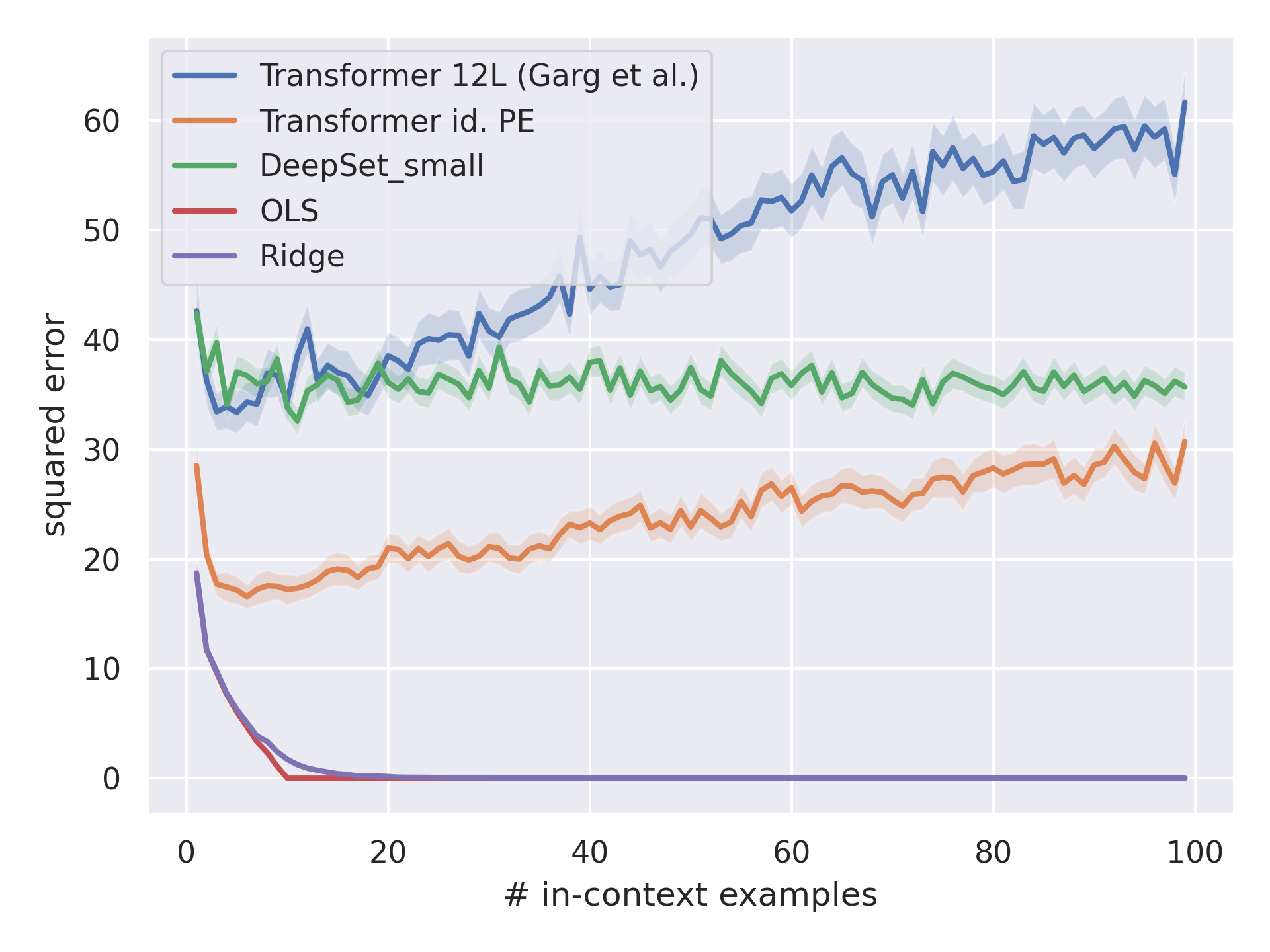}
        \label{fig:label_noise=True_mu=4}
    }
    \vspace{-0.1in}
    \caption{ICL performance under various distribution shifts. $\sigma=0$ for the first row and $\sigma=1$ for the second row.}
    \label{fig:ICL_performance}
    \vspace{-0.15in}
\end{figure}

We first show the results of DeepSet and Transformer across the aforementioned distribution shifts in Fig.~\ref{fig:ICL_performance}.
We use a $d$ of $10$ so that it requires at least $10$ in-context examples to find the underlying solution $w$.
It can be found that, when $\mu=0$ (Fig.~\ref{fig:ICL_performance}(a,d)), both DeepSet and Transformer exhibit great ICL ability and perform similarly as OLS and ridge regression within the context window $k$ of $50$. However, when $k$ grows to $100$,  the Transformer slightly loses the ICL capability while DeepSet remains on par with OLS and ridge regression.

When $\mu$ grows to $2$ (Fig.~\ref{fig:ICL_performance}(b,e)), DeepSet loses the ICL ability, while Transformer performs more similarly to OLS and ridge regression. Nevertheless, as $k$ grows, the performance of the Transformer will be decreased similarly to DeepSet.
When $\mu$ grows larger to $4$ (Fig.~\ref{fig:ICL_performance}(c,f)), the Transformer completely underperforms DeepSet among all $k$ while DeepSet remains as stable as $\mu=2$.

Interestingly, across all $\mu$ and $\sigma$, the performance of DeepSet is consistently stable for all $k\geq 0$, despite its simple architecture and relatively small parameters.

\subsection{ICL Invariance}
To better understand the phenomenon, we further consider the differences between a DeepSet and the Transformer architecture. Apparently, the fundamental difference between the two architectures lies in how they model the input demonstrations.
In fact, the objective of ICL is essentially to learn to infer the label of the query example based on a \textit{set} of demonstrations. In other words, given the same query, the order of demonstrations should not affect the output of the model.

\begin{definition}[ICL Invariance]
    Consider the same data generation process as in Sec.~\ref{sec:icl_lin}, denote the prompt prefix space as $\gP^i$, a model $M:\gP^i\rightarrow\R$ commits to ICL invariance iff. $M(p^i)=M(\pi(p^i)), \forall i$, for any permutation on demonstrations $\pi(p^i):=(x_{\pi(1)},y_{\pi(1)},...,x_{\pi(i)},y_{\pi(i)},x_{i+1})$, where $\pi(i)$ denotes the permuted index.
\end{definition}

By definition, we know the optimal solution $M$ to Eq.~\ref{eq:icl_loss} satisfies the ICL Invariance, and so as DeepSet.
If ICL invariance is crucial for ICL OOD generalizable models, then equipping the Transformer with ICL invariance can further improve its generalizability under the ICL distribution shifts.
One of the key ingredients in the Transformer that breaks ICL invariance is the positional encoding. Therefore, we further evaluate a modified Transformer with identical positional encodings, which commits to ICL invariance better than the original Transformer architecture.

The results are given in Fig.~\ref{fig:ICL_performance}. It can be found that the Transformer with identical positional encoding also exhibits the ICL ability and can perform even better than DeepSet under $\mu=0$ across different context window sizes. Moreover, even with distribution shifts on the demonstration inputs and labels, the Transformer with identical positional encoding can retain the great ICL capability and perform competitively as OLS and ridge regression. When $\mu$ grows, the Transformer with identical positional encoding still performs better than DeepSet and the original Transformer architecture.
The superior ICL performance of the Transformer with identical positional encoding across various distribution shifts serves as strong evidence for the importance of ICL invariance.

\section{Conclusions}
Through the study of ICL linear regression with various distribution shifts, we identify an important property for OOD generalizable ICL, i.e., ICL invariance. We show simply modifying the positional encoding in the Transformer to better commit to ICL invariance can retain great ICL capability across multiple ICL distribution shifts.
As auto-regressive models will break the ICL invariance explicitly or implicitly~\citep{still_learn_pe}, a promising future direction is to incorporate ICL invariance into pre-trained LLMs.
Besides, it is also interesting to find what causes the better OOD generalizability in ICL than DeepSet in the Transformer when both of them commit to ICL invariance.

\bibliographystyle{references}
\bibliography{references,llm}

\end{document}